\documentclass{article}
\usepackage{spconf,amsmath,graphicx}
\usepackage{epstopdf}
\usepackage{amssymb}
\usepackage{dsfont}

\usepackage{multirow}
\usepackage{hhline}

\usepackage[dvipsnames]{xcolor}
\def\etal{\emph{et al.~}}
\usepackage[normalem]{ulem}

\title{Using CycleGANs for effectively reducing image variability across OCT devices and improving retinal fluid segmentation}

\makeatletter
\def\@name{ \emph{$^{\star}$Philipp Seeb{\"o}ck$^{1,2}$, $^{\star}$David Romo-Bucheli$^{1}$, Sebastian Waldstein$^{1}$, Hrvoje Bogunovic$^{1}$},  \\ \emph{Jos{\'e} Ignacio Orlando$^{1}$,  Bianca S. Gerendas$^{1}$, Georg Langs$^{1,2}$, Ursula Schmidt-Erfurth$^{1}$}\thanks{The financial support by the Christian Doppler Research Association, the Austrian Federal Ministry for Digital and Economic Affairs and the National Foundation for Research, Technology and Development and by the Austrian Science Fund (FWF I2714-B31) is gratefully acknowledged. A TitanX used for this research was donated by the NVIDIA Corporation.}\thanks{\textbf{$^{\star}$ Contributed equally (order was defined by flipping a coin)}}}
\makeatother

\address{\\
	$^{1}$ Christian Doppler Laboratory for Ophthalmic Image Analysis, Department of Ophthalmology,\\ Medical University of Vienna, Austria\\
	$^{2}$ Computational Imaging Research Lab, Department of Biomedical Imaging and Image-guided Therapy,\\ Medical University Vienna, Austria
}

\begin{document}
\maketitle
\begin{abstract}
Optical coherence tomography (OCT) has become the most important imaging modality in ophthalmology. A substantial amount of research has recently been devoted to the development of machine learning (ML) models for the identification and quantification of pathological features in OCT images. Among the several sources of variability the ML models have to deal with, a major factor is the acquisition device, which can limit the ML model's generalizability.
In this paper, we propose to reduce the image variability across different OCT devices (Spectralis and Cirrus) by using CycleGAN, an unsupervised unpaired image transformation algorithm. The usefulness of this approach is evaluated in the setting of retinal fluid segmentation, namely intraretinal cystoid fluid (IRC) and subretinal fluid (SRF).
First, we train a segmentation model on images acquired with a source OCT device. Then we evaluate the model on (1) source, (2) target and (3) transformed versions of the target OCT images. The presented transformation strategy shows an $F_1$ score of $0.4$ ($0.51$) for IRC (SRF) segmentations. Compared with traditional transformation approaches, this means an $F_1$ score gain of $0.2$ ($0.12$).
\end{abstract}

\begin{keywords}
covariate shift, optical coherence tomography, generative adversarial networks, image segmentation 
\end{keywords}

\section{INTRODUCTION}
\label{sec:intro}
Automated methods are being developed for medical image analysis to solve clinical tasks, such as segmentation of anatomical structures or classification of normal/pathological cases. In the last decade, machine learning (ML) techniques have been showing an increasing ability to solve these problems. In particular, deep neural networks (DNN) have obtained impressive results in several medical imaging tasks \cite{litjens2017survey}. Nevertheless, these models can still under-perform in ``deployment'' datasets. One important factor explaining the difference in the performance of ML models is covariate shift \cite{Storkey_2009}. Covariate shift is a phenomenon observed ``when the data is generated according to a model $P(y|x)$ and where the distribution $P(x)$ changes between training and test scenarios''\cite{Storkey_2009}. 

Among the different medical imaging modalities, optical coherence tomography (OCT) provides high-resolution 3D volumes of the retina, is non-invasive and the most important diagnostic modality in ophthalmology. A single OCT volume is composed of multiple cross-sectional images known as B-scans. Current treatment and diagnosis guidelines rely on the examination of these B-scans to inform clinical decisions \cite{Fujimoto_2016}. Several machine learning techniques have recently been proposed help identify and quantify retinal pathological features\cite{Schmidt_Erfurth2018}. Unfortunately, ML models are often susceptible to covariate shift when training on data from a specific vendor, meaning that differences in the intensity distribution, resolution and noise level can affect the generalization ability of the model. For instance, significant differences between OCT acquisition devices in retinal fluid segmentation performance~\cite{Schlegl_2018} or automated layer thickness measurements \cite{terry_2016} have been reported. A common strategy to deal with image variability across multiple devices is training vendor-specific models. For instance, De Fauw~\etal~\cite{de2018clinically} proposed a two-stage deep learning approach for diagnosis and referral in retinal disease, using scans from two OCT vendors. Time-consuming manual annotations are needed for each vendor to successfully (re-)train the segmentation model and hence achieve device independence. In contrast, our approach does not require additional manual annotations.

In this work, we present a strategy based on CycleGANs~\cite{cyclegan_2018} to reduce the covariate shift between different OCT acquisition devices, and consequently to improve the robustness of ML fluid segmentation models. 


\section{MATERIALS}
\label{sec:materials}
\begin{figure}[ht]
	\includegraphics[width=8.5cm]{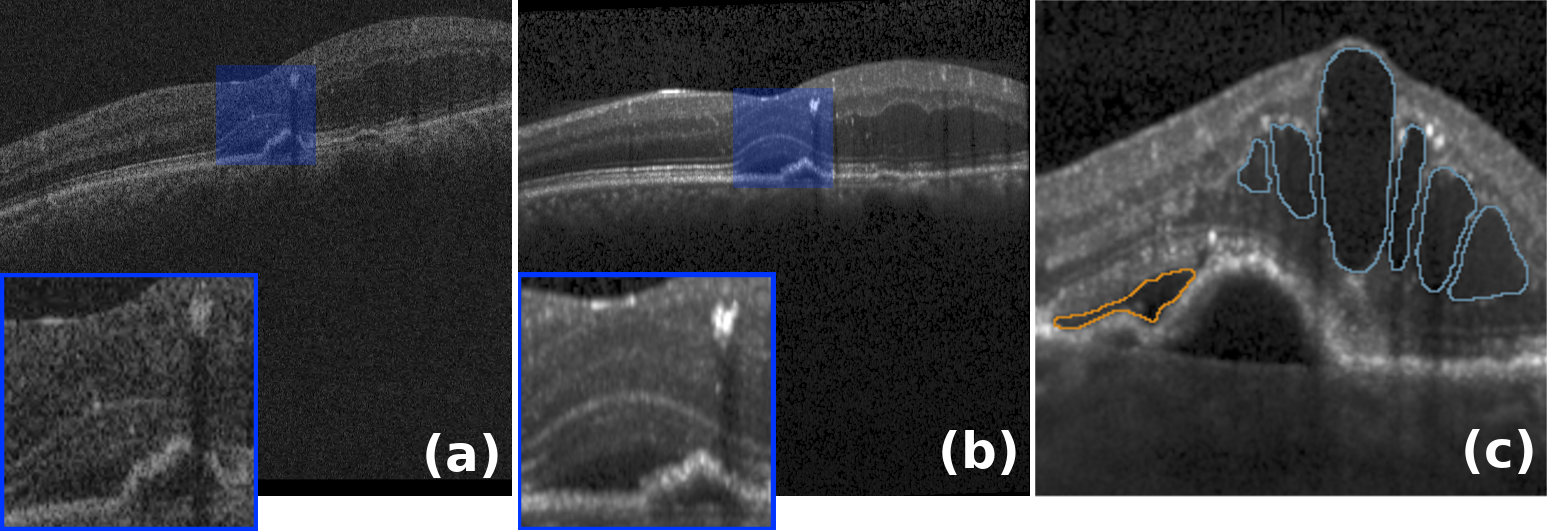}
	\caption{Spectralis B-scan \textbf{(a)} and Cirrus B-scan \textbf{(b)} with a corresponding close-up of the retinal layers. Both B-scans were acquired from the same patient at approximately the same time and retinal location. The difference in intensity values and noise level is noticeable. \textbf{(c)} Spectralis B-scan region with manual delineation of IRC (blue) and SRF (orange).}
	\label{fig:oct_vendors}
\end{figure}

The OCT volumes used in this work were either acquired using Spectralis OCT instruments (Heidelberg Engineering, GER) with voxel dimensions of $496 \times 512 \times 49$, or Cirrus HD-OCT (Carl Zeiss Meditec, Dublin, CA, USA) with voxel dimensions of $1024 \times 512 \times 128$ or $1024 \times 200 \times 200$. For both instruments, the volumes were centered at the fovea of the retina and covered a physical volume of approximately~$2\mu m  \times 6\mu m \times 6 \mu m$. Besides different voxel resolutions within B-scans, Spectralis OCT volumes contain a lower number of B-scans ($49$) than Cirrus ($128$). At the same time, due to B-scan averaging performed by Spectralis devices\footnote{The Spectralis OCT scans with $49$ B-scans are usually acquired with a default of $16$ averaged frames per B-scan.}, which is not conducted in Cirrus, the signal-to-noise ratio (SNR) in Spectralis OCTs is usually better and retinal structures are easier to identify both for human observers and automated methods~\cite{smretschnig2010cirrus,Schlegl_2018}. All Cirrus OCTs are resampled to obtain a $496 \times 512 \times 49$ volume using nearest-neighbor interpolation. A visual comparison between a Cirrus and Spectralis B-scan is presented in Figure~\ref{fig:oct_vendors}. 

\subsection{Retinal cross-domain dataset}
\label{secsec:cross-domain_dataset}
A total set of $1,179$ OCT volumes ($57,771$ B-scans after resampling Cirrus OCTs) comprised the retinal cross-domain dataset. From these, $587$ ($592$) OCT volumes were acquired with a Spectralis (Cirrus) device. Each Spectralis (Cirrus) OCT volume was associated with a retinal disease: $195$ ($192$) retinas with age-related macular degeneration (AMD), $197$ ($196$) with retinal vein occlusion (RVO) and $200$ ($199$) with diabetic macular edema (DME).

\subsection{Retinal fluid-segmentation dataset}
\label{secsec:ret_fluid_dataset}
The retinal fluid-segmentation dataset consisted of $228$ OCT volumes  with a total of $11,172$ B-scans (66 Heidelberg Spectralis and 162 Zeiss Cirrus volumes). The Spectralis (Cirrus) set contained $59 (62)$ pathological retinas with AMD and $7 (100)$ with RVO. Manual pixel-wise annotations of the existing two different retinal types of fluid, intraretinal cystoid fluid (IRC) and subretinal fluid (SRF), were used for all 228 volumes, as shown in the far right panel of Figure~\ref{fig:oct_vendors}. The annotation was performed by experienced graders of the Vienna Reading Center, supervised by retinal expert ophthalmologists following a standardized annotation protocol.




\section{METHODS}
\label{sec:methods}

In this section we present the components used in our experiments. Traditional and unpaired image transformation algorithms were used to transform images. The segmentation model was used to evaluate the effectiveness of transformations.

\subsection{Baseline transformation algorithms}
\label{subsec:common_trans}
Many image preprocessing techniques are used to enhance and reduce the variability in OCT images. In this work, we followed the pre-processing pipeline in \cite{tennakoon2018retinal} to define two suitable transformations from Cirrus to Spectralis OCT volumes.
%
The first transformation strategy (\emph{$T_1$}) uses an initial median filtering operation (with $3 \times 3$ kernel size) followed by a second median filtering operation across B-scans, with a $1 \times 1 \times 3$ kernel size.
The second transformation (\emph{$T_2$}) performs an initial histogram matching step using a random Spectralis OCT volume as a template and subsequently, the same filtering operations as described for $T_{1}$ are applied.


\subsection{Unsupervised unpaired transformation algorithm}
\label{subsec:cgan_trans}
Cycle generative adversarial networks (CycleGANs) allow a suitable transformation function between different image domains to be discovered in an unsupervised way\cite{cyclegan_2018}. The CycleGAN uses two discriminator-generator pairs ($G_1$-$D_1$, $G_2$-$D_2$) which are implemented as deep neural networks. $G_1$ is fed with an image from the source domain and transforms it into the target domain. Then, $G_2$ transforms the image back from the target to the source domain. $D_1$ ($D_2$) is trained to distinguish between real samples from the target (source) domain and the transformed images. Additionally, the cycle consistency and the identity loss \cite{cyclegan_2018} are used as regularizers to avoid trivial solutions and obtain a meaningful mapping.

\subsection{Segmentation model}
In this work, we performed the segmentation task in 2D, B-scan wise. Given an OCT B-scan $I \in \mathbb{R}^{a \times b}$ and a corresponding target label map $S \in \mathbb{R}^{a \times b}$, the segmentation model aims at finding the function $h: I \rightarrow S$.
The U-Net~\cite{ronneberger2015u} was used as the segmentation model, which is composed of an encoding and a decoding part. The encoder consists of convolution blocks followed by max-pooling layers, which contracts the input and uses the context for segmentation. The decoder counterpart performs up-sampling operations followed by convolution blocks to enable precise localization. 

\section{EXPERIMENTAL SETUP}
\label{sec:setup}
In our experiments we evaluated if the image transformation algorithms improved the generalization ability of a trained model on a new unseen domain.

\subsection{Unsupervised image transformation details}
For the unsupervised unpaired image transformation task, the CycleGAN models were trained by using the cross-domain dataset (Section~\ref{secsec:cross-domain_dataset}). For each vendor, the data was randomly split into training ($90\%$) and validation sets ($10\%$), with no patient overlap between these two sets.\\
Four different CycleGAN models were trained with image patches cropped within B-scans in sizes of $64$, $128$, $256$, and $460$. For each configuration, the models were trained for $20$ epochs with a batch size of $1$. The generator and discriminator were stored at the end of each epoch.
A model selection procedure found the best performing model out of $20$ to compensate for the instability of the adversarial loss during training. To select one generator model, we used the pool of $20$ generators to create $20$ transformed validation sets. For each transformed set, we applied all $20$ discriminators and used the maximum adversarial loss as the selection score. The model with the lowest selection score was chosen. This resulted in four pairs of generator models (Cirrus-to-Spectralis, Spectralis-to-Cirrus), one for each patch size configuration (\textit{CGAN-64}, \textit{CGAN-128}, \textit{CGAN-256}, \textit{CGAN-460}).


\subsection{Segmentation model details}
Following the original U-Net architecture, we used five levels of depth, the number of channels going down from $64$ in the first to $1024$ in the bottleneck layer. Each convolutional block consisted of two $3 \times 3$ convolutions, each followed by a batch-normalization layer and a rectified linear unit (ReLU). While $2 \times 2$ max-pooling was used for downsampling, upsampling was performed using nearest-neighbor interpolation.\\
We used the negative log-likelihood loss in all our segmentation experiments. Kaiming initialization~\cite{he2015delving}, Adam optimization~\cite{kingma2014adam}, and a learning rate of $0.0001$ which was decreased by half every $15$ epochs were used. We trained our networks for $80$ epochs and selected the model with the best average $F_1$-score on the validation set. We used a random separation of training ($70\%$), validation ($10\%$) and test set ($20\%$) on patient-distinct basis for all experiments.

\begin{figure}[t]
	\includegraphics[width=8.5cm, height = 7.8cm]{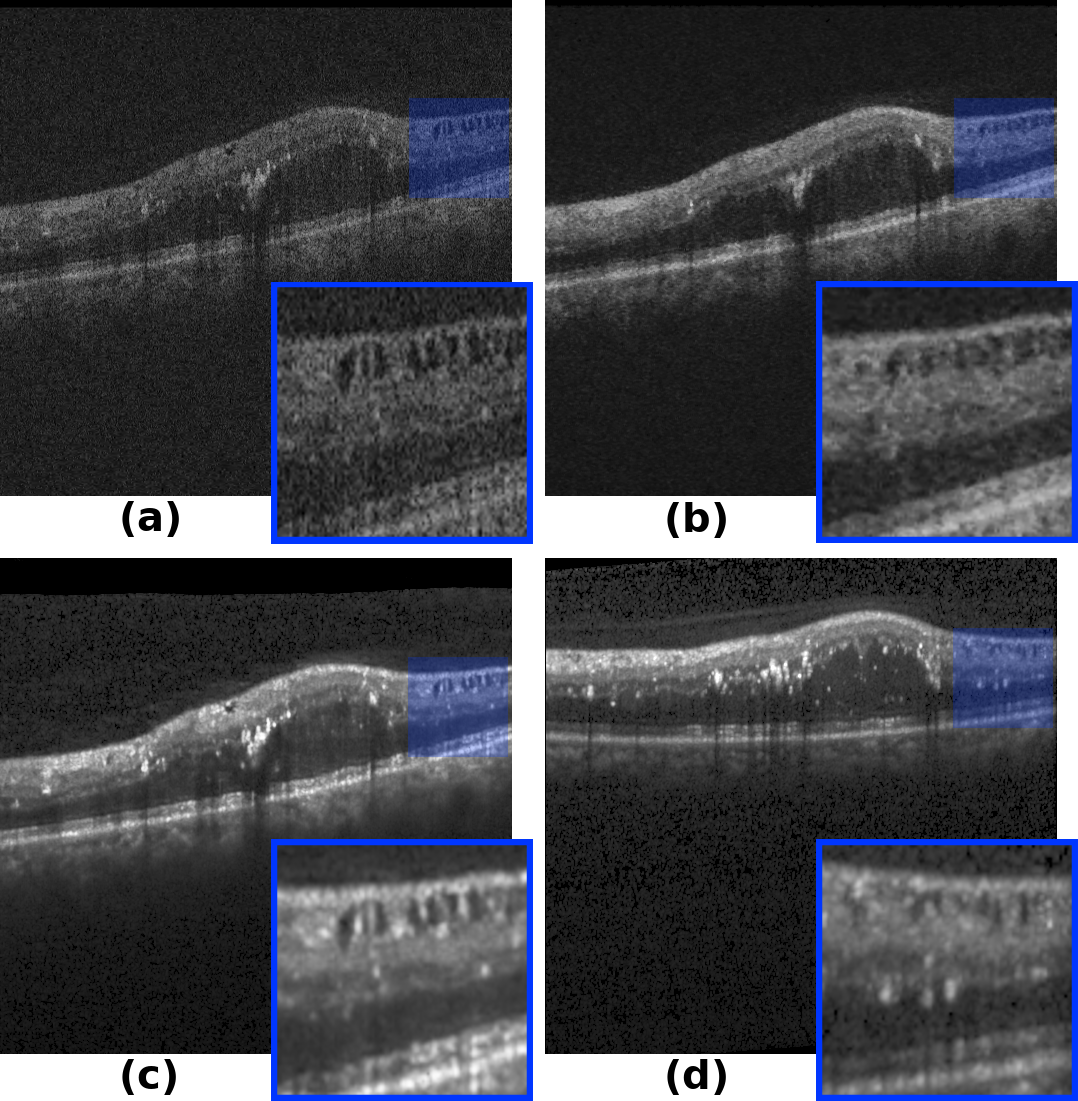}
	\caption{Qualitative results of the image transformation algorithms. An original Cirrus OCT B-scan \textbf{(a)} was transformed to the Spectralis domain using $T_2$ \textbf{(b)}, and \textit{CGAN-$460$} \textbf{(c)}.  The corresponding original Spectralis B-scan \textbf{(d)} acquired from the same patient at approximately the same time and retinal location is also shown. The image generated in \textbf{(c)} has intensity values and an image noise level similar to those observed in the original Spectralis image \textbf{(d)}.
	}
	
	\label{fig:qualitative_preprocessing}
\end{figure}

\subsection{Model generalization to a different domain}
The good SNR of Spectralis OCT volumes (described in Section \ref{sec:materials}) facilitates the manual annotation of retinal structures. Hence, a typical use-case of the presented strategy would be to collect manual annotations and train a segmentation model on Spectralis scans but apply the model both on Spectralis and transformed Cirrus scans.
%
Accordingly, this experiment involved training the segmentation model on a source dataset (Spectralis) and evaluating it on a target dataset (Cirrus). We applied common transformation techniques (Section~\ref{subsec:common_trans}) as well as CycleGAN transformations (Section~\ref{subsec:cgan_trans}) to convert OCTs from the Cirrus into the Spectralis domain. The underlying assumption is that a higher segmentation performance indicates a more effective transformation. 
We used a segmentation model, trained on the target domain, as an upper bound for our performance.


\section{RESULTS}
\label{sec:results}
The transformation algorithms were used to generate different versions of the target dataset (Cirrus). Qualitative results are illustrated in Figure~\ref{fig:qualitative_preprocessing}.
%
%
Initial evaluation of the Spectralis segmentation model on the Spectralis test set yields a precision, recall and $F_1$-score for the IRC (SRF) of $0.73$ ($0.75$), $0.64$ ($0.74$), $0.69$ ($0.75$).
The quantitative results of the Spectralis model on the Cirrus test set, comparing different transformation strategies, as well as results of the Cirrus model~(\textit{*Cirrus-on-Cirrus}) are shown in Table~\ref{tab:results_segmentation_transformation}. 

\begin{table}[ht] \footnotesize
	
	\begin{center}
		\begin{tabular}{l|rrr|rrr}
			\multirow{2}{*}{Transformation} & \multicolumn{3}{c|}{IRC} & \multicolumn{3}{c}{SRF} \\
			& Pr & Rec & F1 & Pr & Rec & F1  \\
			\hhline{=======}
			None & \textbf{0.73} & 0.01 & 0.01 & 0.39 & 0.01 & 0.01 \\
			$T1$ & 0.73 & 0.11 & 0.19 & 0.38 & 0.23 & 0.29 \\
			$T2$ & 0.16 & 0.27 & 0.20 & 0.44 & 0.38 & 0.40 \\
			CGAN-64  & 0.25 & 0.18 & 0.21 & 0.18 & 0.17 & 0.17 \\
			CGAN-128 & 0.39 & 0.27 & 0.31 & 0.49 & 0.35 & 0.41 \\
			CGAN-256 & 0.27 & 0.38 & 0.31 & 0.44 & 0.47 & 0.45 \\
			CGAN-460 & 0.60 & \textbf{0.30} & \textbf{0.40} & \textbf{0.60} & \textbf{0.45} & \textbf{0.52} \\
			\hline
			\textit{*Cirrus-on-Cirrus} & \textit{0.57} & \textit{0.64} & \textit{0.58} & \textit{0.83} & \textit{0.43} & \textit{0.57} \\
			\hline
		\end{tabular}
		\caption{Performance of the Spectralis segmentation model applied on different ``versions'' of the Cirrus test set. The last row corresponds to the performance of the Cirrus model applied on the non-transformed Cirrus test set. (Pr=Precison, Rec=Recall)}
		\label{tab:results_segmentation_transformation}
		\vspace{-2pt}
	\end{center}
\end{table}

A few insights can be extracted from Table~\ref{tab:results_segmentation_transformation}. First, the Spectralis DL segmentation model performance drops dramatically when directly applied on the Cirrus target dataset. The effect of the covariate shift in the model is evident. Secondly, the histogram-based combined with the filtering operations in $T_2$ perform better than no transformation at all. Thirdly, the best cross-vendor segmentation performance is obtained with a CycleGAN-based transformation strategy (trained with a $460 \times 460$ patch size). The performance was certainly lower than those observed in Spectralis B-scans, but was closer to the performance a Cirrus model would obtain on the Cirrus B-scans. More precisely, while a direct application of the segmentation model on non-transformed data yields a $F_1$-score close to $0$, the CGAN-$460$ approach yielded a $0.40$ ($0.52$) $F_1$-score for IRC (SRF) segmentations. With respect to the best traditional transformation technique, a $0.2$ ($0.12$) $F_1$-score gain was observed.

\section{CONCLUSIONS}
\label{sec:conclusions}
We present a CycleGAN-based strategy to reduce the image variability across OCT acquisition devices. The results show that the transformation algorithm improves the performance of a fluid segmentation model on a target dataset, thus effectively reducing the covariate shift (i.e difference between the target and source datasets). This finding is relevant as automated fluid segmentation could potentially be part of routine diagnostic workflows and affect therapy of millions of patients. In this scenario, the presented approach would reduce device dependency of ML algorithms and therefore allow more clinicians to use them on their specific OCT device. Recently, it has been shown that the CycleGAN algorithm could induce misleading diagnosis in medical images, and hence the transformed images are not recommended for direct visualization and interpretation by a clinician~\cite{cohen_2018}. This is prominent when the distribution of pathologies differs between the two domain sets. We carefully selected OCTs from the two devices, containing similar distribution of pathological and healthy retina. Therefore, we expect most of the difference to arise from image intensity properties and the SNR of the devices. Furthermore, in this work we use CycleGAN only as a pre-processing step for facilitating automated segmentation. 
Future work will be focused on evaluating the presented strategy in other ML tasks.


\bibliographystyle{IEEEbib}
\bibliography{cirrus_spectralis}

\begin{thebibliography}{10}

\bibitem{litjens2017survey}
Geert Litjens, Thijs Kooi, Babak~Ehteshami Bejnordi, et~al.,
\newblock ``A survey on deep learning in medical image analysis,''
\newblock {\em MED IMAGE ANAL}, vol. 42, pp. 60--88, 2017.

\bibitem{Storkey_2009}
Amos~J. Storkey,
\newblock {\em Dataset Shift in Machine Learning}, chapter When training and
  test sets are different: characterising learning transfer, pp. 3--28,
\newblock The MIT Press, 2009.

\bibitem{Fujimoto_2016}
James Fujimoto and Eric Swanson,
\newblock ``The development, commercialization, and impact of optical coherence
  tomography,''
\newblock {\em INVEST OPHTH VIS SCI}, vol. 57, no. 9, pp. OCT1--OCT13, 2016.

\bibitem{Schmidt_Erfurth2018}
Ursula Schmidt-Erfurth, Amir Sadeghipour, Bianca~S. Gerendas, et~al.,
\newblock ``Artificial intelligence in retina,''
\newblock {\em PROG RETIN EYE RES}, 2018.

\bibitem{Schlegl_2018}
Thomas Schlegl, Sebastian~M. Waldstein, Hrvoje Bogunovic, et~al.,
\newblock ``Fully automated detection and quantification of macular fluid in
  oct using deep learning,''
\newblock {\em Ophthalmology}, vol. 125, no. 4, pp. 549 -- 558, 2018.

\bibitem{terry_2016}
Louise Terry, Nicola Cassels, Kelly Lu, et~al.,
\newblock ``Automated retinal layer segmentation using spectral domain optical
  coherence tomography: evaluation of inter-session repeatability and agreement
  between devices,''
\newblock {\em PloS one}, vol. 11, no. 9, pp. e0162001, 2016.

\bibitem{de2018clinically}
Jeffrey De~Fauw, Joseph~R Ledsam, Romera-Paredes, et~al.,
\newblock ``Clinically applicable deep learning for diagnosis and referral in
  retinal disease,''
\newblock {\em NATURE MED}, vol. 24, no. 9, pp. 1342--1350, 2018.

\bibitem{cyclegan_2018}
J.~Zhu, T.~Park, P.~Isola, et~al.,
\newblock ``Unpaired image-to-image translation using cycle-consistent
  adversarial networks,''
\newblock in {\em Proc. of IEEE ICCV}, 2017, pp. 2242--2251.

\bibitem{smretschnig2010cirrus}
Eva Smretschnig, Ilse Krebs, Sarah Moussa, et~al.,
\newblock ``Cirrus oct versus spectralis oct: differences in segmentation in
  fibrovascular pigment epithelial detachment,''
\newblock {\em GRAEF ARCH CLIN EXP}, pp. 1693--1698, 2010.

\bibitem{tennakoon2018retinal}
Ruwan Tennakoon, Amirali~K Gostar, Reza Hoseinnezhad, et~al.,
\newblock ``Retinal fluid segmentation in oct images using adversarial loss
  based convolutional neural networks,''
\newblock in {\em Proc. of the IEEE 15th ISBI}. IEEE, 2018, pp. 1436--1440.

\bibitem{ronneberger2015u}
Olaf Ronneberger, Philipp Fischer, and Thomas Brox,
\newblock ``U-net: Convolutional networks for biomedical image segmentation,''
\newblock in {\em Proc. of MICCAI}. Springer, 2015, pp. 234--241.

\bibitem{he2015delving}
Kaiming He, Xiangyu Zhang, Shaoqing Ren, et~al.,
\newblock ``Delving deep into rectifiers: Surpassing human-level performance on
  imagenet classification,''
\newblock in {\em Proc. of IEEE ICCV}, 2015, pp. 1026--1034.

\bibitem{kingma2014adam}
Diederik~P Kingma and Jimmy Ba,
\newblock ``Adam: A method for stochastic optimization,''
\newblock {\em arXiv preprint arXiv:1412.6980}, 2014.

\bibitem{cohen_2018}
Joseph~Paul Cohen, Margaux Luck, and Sina Honari,
\newblock ``Distribution matching losses can hallucinate features in medical
  image translation,''
\newblock in {\em Proc. of MICCAI}, 2018, pp. 529--536.

\end{thebibliography}

\end{document}